\documentclass[preprint,12pt,authoryear]{elsarticle}

\usepackage{amssymb}
\usepackage{subfigure}
\usepackage{makeidx}  % allows for indexgeneration
\usepackage{multirow}
\usepackage{multicol}
\usepackage{footmisc}
\usepackage{booktabs}
\usepackage{rotating}
\usepackage{amsmath,amsfonts,amssymb}
\usepackage{bm}
\usepackage{url}
\usepackage[svgnames,dvipsnames]{xcolor}
\usepackage{cprotect}
\usepackage[english]{babel}
\usepackage{float}
\usepackage{adjustbox}
\usepackage{tabularx}
\usepackage{subfig}
\usepackage[ruled,vlined,linesnumbered,titlenumbered]{algorithm2e}
\usepackage{algcompatible}
\usepackage{times}
\usepackage{soul}
\usepackage{verbatim}

\newcolumntype{P}[1]{>{\centering\arraybackslash}p{#1}}
\newcolumntype{M}[1]{>{\centering\arraybackslash}m{#1}}
\usepackage{enumitem}
\usepackage{marginnote}
\usepackage{soul}
\usepackage{listings}

\journal{}

\newcolumntype{L}[1]{>{\raggedright\let\newline\\\arraybackslash\hspace{0pt}}m{#1}}
\newcolumntype{C}[1]{>{\centering\let\newline\\\arraybackslash\hspace{0pt}}m{#1}}
\newcolumntype{R}[1]{>{\raggedleft\let\newline\\\arraybackslash\hspace{0pt}}m{#1}}

\newcounter{inlineenum}
\renewcommand{\theinlineenum}{\alph{inlineenum}}

\begin{document}

\begin{frontmatter}



\title{Spiking Neural Networks and Online Learning: An Overview and Perspectives}

\author[label1]{Jesus L. Lobo\corref{cor1}}
\author[label1,label2,label3]{Javier Del Ser} 
\author[label4,label5]{Albert Bifet}
\author[label6]{Nikola Kasabov}

\address[label1]{TECNALIA, 48160 Derio, Spain.}
\address[label2]{Basque Center for Applied Mathematics (BCAM), 48009 Bilbao, Spain}
\address[label3]{University of the Basque Country UPV/EHU, 48013 Bilbao, Spain}
\address[label4]{T\'{e}l\'{e}com ParisTech, Par\'{\i}s, C201-2 France}
\address[label5]{University of Waikato, Hamilton, New Zealand}
\address[label6]{Auckland University of Technology (AUT), Auckland, New Zealand}
\cortext[cor1]{Corresponding author: jesus.lopez@tecnalia.com (Jesus L. Lobo). TECNALIA. P. Tecnologico Bizkaia, Ed. 700, 48160 Derio, Spain. Tl: +34 946 430 50. Fax: +34 901 760 009.}

\begin{abstract}
\small{Applications that generate huge amounts of data in the form of fast streams are becoming increasingly prevalent, being therefore necessary to learn in an online manner. These conditions usually impose memory and processing time restrictions, and they often turn into evolving environments where a change may affect the input data distribution. Such a change causes that predictive models trained over these stream data become obsolete and do not adapt suitably to new distributions. Specially in these non-stationary scenarios, there is a pressing need for new algorithms that adapt to these changes as fast as possible, while maintaining good performance scores. Unfortunately, most off-the-shelf classification models need to be retrained if they are used in changing environments, and fail to scale properly. Spiking Neural Networks have revealed themselves as one of the most successful approaches to model the behavior and learning potential of the brain, and exploit them to undertake practical online learning tasks. Besides, some specific flavors of Spiking Neural Networks can overcome the necessity of retraining after a drift occurs. This work intends to merge both fields by serving as a comprehensive overview, motivating further developments that embrace Spiking Neural Networks for online learning scenarios, and being a friendly entry point for non-experts.}
\end{abstract}


\end{frontmatter}


\section{Introduction}

The term \textit{Big Data} has gained progressive momentum during the last decade, due to the feasibility of collecting data from almost any source and analyzing to achieve data-based insights that enable cost and time reductions, new product developments, optimized offerings, or smart decision making, among others profits. In these \textit{Big Data} scenarios, some characteristics may play a relevant role: it is not feasible to store the whole dataset, traditional algorithms cannot handle data produced at high rates, and changes in data distribution may occur during learning process. An increasing number of applications are based on these training data continuously available (\textit{stream learning}), and applied to real scenarios, such as mobile phones, sensor networks, industrial process controls and intelligent user interfaces, among others \citep*{vzliobaite2016overview}. Some of these applications produce non-stationary data streams which are becoming increasingly prevalent, and where the process generating the data may change over time, producing changes in the patterns to be modeled (\textit{concept drift}). This causes that predictive models trained over these streaming data become obsolete and do not adapt suitably to the new distribution. Especially in Online Learning (OL) scenarios, where only a single sample is provided to the learning algorithm at every time instant, there is a pressing need for new algorithms that adapt to these changes as fast as possible, while maintaining good performance scores. OL in the presence of concept drift has been a very hot topic during the last few years \citep*{gama2014survey,webb2016characterizing}, and still remains under active debate in the community because of its numerous open challenges \citep*{krawczyk2017ensemble,losing2018incremental}. The data mining community prefers to refer to OL in the presence of concept drift as \textit{data stream mining} \citep*{DeFrancisciMorales2016,MOA-Book-2018}.

Many algorithms have been developed for stream learning based on Machine Learning (ML) techniques. Unfortunately, most off-the-shelf models need to be retrained if they are used in an evolving environment, and fail to scale properly due to their learning algorithm. Artificial Neural Networks (ANNs) have been used in the last years to deal with these fast evolving information flows. In essence, they are a biologically inspired paradigm that mimics the process through which the brain acquires and processes sensory information. One of their most biologically plausible neuron models is a key ingredient of the so-called Spiking Neural Network (SNN) \citep*{gerstner2002spiking}, a popular an reputed model for its capacity to capture informational dynamics observed among real biological neurons, and to represent and integrate several information dimensions (e.g. time, space, frequency, phase, and to deal with large volumes of data) into a single model. The theory behind SNNs is currently mostly accepted to describe realistic \emph{brain-like} information processing, which in addition eases their implementation on super-fast and reliable hardware platforms. 

Considered nowadays as the third generation of ANNs, the advent of SNN was propelled by the need for a better understanding of the information processing skills of the mammalian brain, for which the community committed itself to the development of more complex biologically connectionist systems. Some SNNs are especially well-known in the OL research community for their ability to learn continuously and incrementally, which account for their continuous adaptability to non-stationary and evolving environments, and also  their capacity to serve as drift detectors \citep*{lobo2018drift}. Besides, they have shown the ability to capture temporal associations between temporal variables in streaming data.

From all the rationale exposed above, the merging of both fields motivates further developments that embrace SNNs for OL scenarios, with emphasis on those requiring concept drift detection and adaptation. This work intends to serve as a suitable entry point in the literature for non-experts in both fields, and a catalyst material for future research efforts invested in this direction; it is organized as follows: while, Section \ref{OL} and Section \ref{SNNs_sec} provide a general introduction, and present challenges and future work for OL scenarios and SNNs respectively, Section \ref{SNNs_OL} delves into the convergence of both fields. Finally, Section \ref{conc} draws the conclusions related to this study. 

\section{Online Learning}\label{OL}

In stream learning, data may arrive in chunks of data (\textit{batch learning}) or in an online manner, i.e., one single sample at a time (OL). In \textit{batch learning} an entirely accessible group of samples (batch) is provided, and the learning algorithm is allowed to scan the batch before building/updating the model. However, in OL only a single sample is provided to the learning algorithm at every time instant, which is incrementally updated every time a new sample arrives. An OL environment imposes different computational constraints when compared to traditional batch learning setting:

\begin{itemize}
	\item each sample is processed only once \textit{on arrival}, and models must be able to process samples sequentially as soon as they are received, without putting the memory space and processing time restrictions at risk;
	\item the processing time of each sample must be small and constant, without exceeding the rate at which new samples arrive;
	\item the algorithm should use only a preallocated and finite amount of memory;
	\item a valid model must be available at every scan of the streams of data; and
	\item the learning algorithm must produce a model that is equivalent to the one that would be built in a batch learning scenario.
\end{itemize}

In \textit{batch learning}, the evaluation procedure of the learning algorithm is determined by the set of samples used for training and testing. The question raised for OL in this regard is how to build a picture of accuracy over time. One of the most used schemes is \textit{test-then-train}, where each sample is used to test the model before it is used for training, and then the accuracy can be incrementally updated. This scheme has the advantage that can be applied when memory is restricted and there is no holdout set for testing, getting the most out of the available dataset. Next, we introduce the problem of OL in the presence of concept drift, probably the most challenging aspect in OL, being a very hot topic research in the last decade \citep*{webb2016characterizing}.

\subsection{Concept Drift}

\begin{figure}[!ht]
	\centering
	\captionsetup{justification=centering}	
	
	\subfigure[Original, virtual ($p(\textbf{x})$ changes, but not $p(y|\textbf{x})$) and real ($p(y|\textbf{x})$ changes) concepts respectively]{\includegraphics[width=0.18\columnwidth]{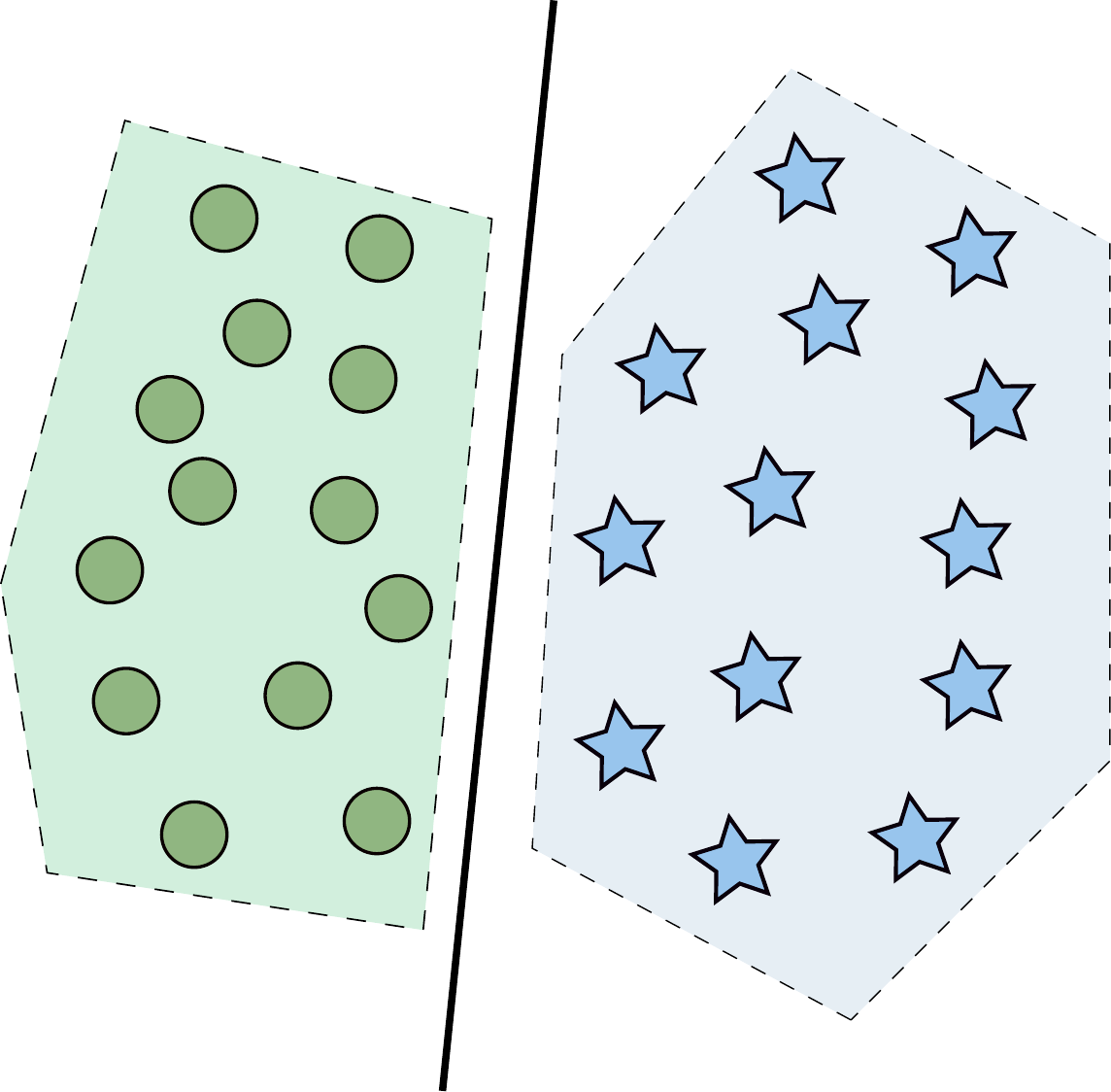} \hfill
		\hspace{1mm}
		\includegraphics[width=0.18\columnwidth]{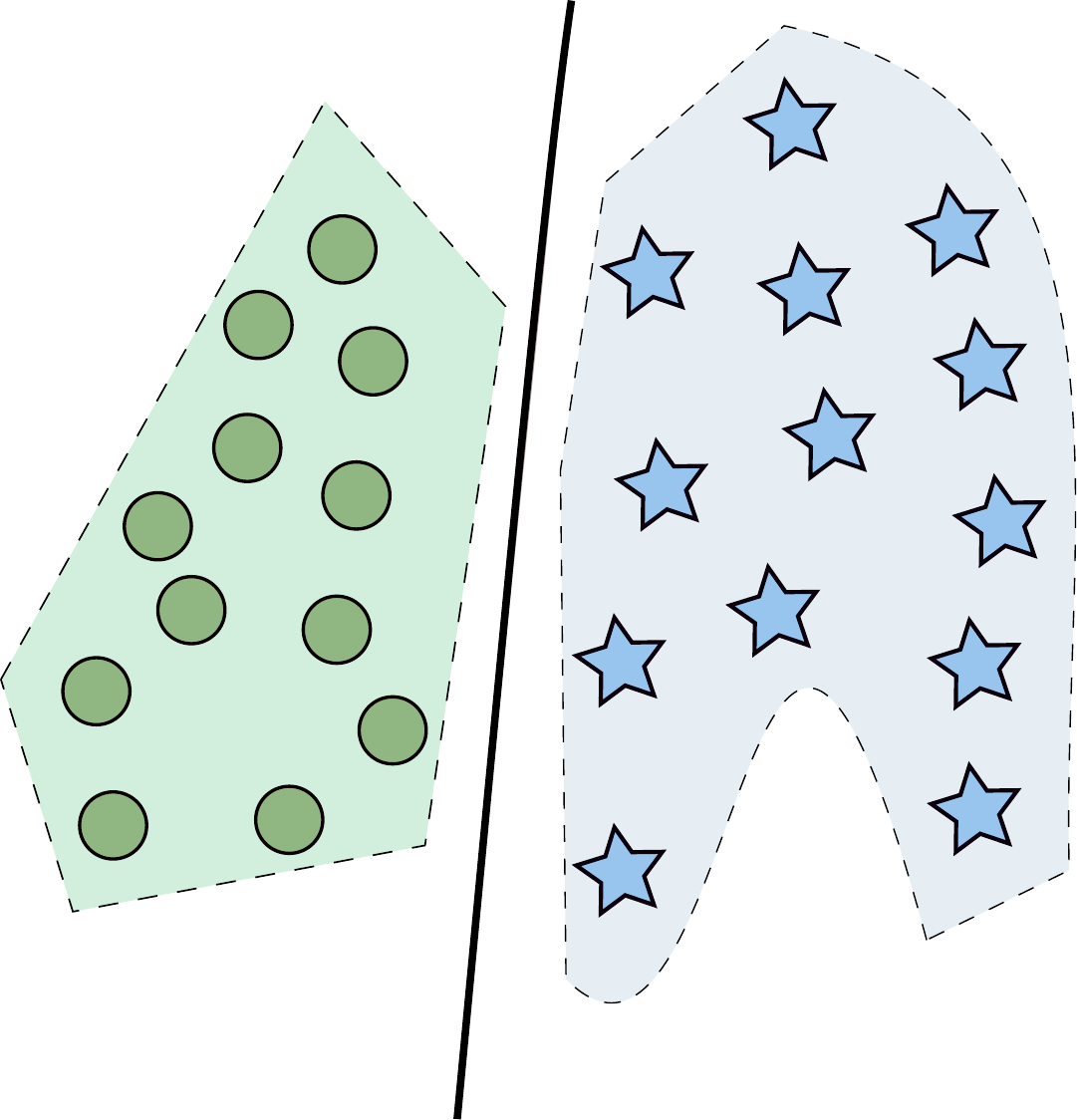} \hfill
		\hspace{1mm}	
		\includegraphics[width=0.18\columnwidth]{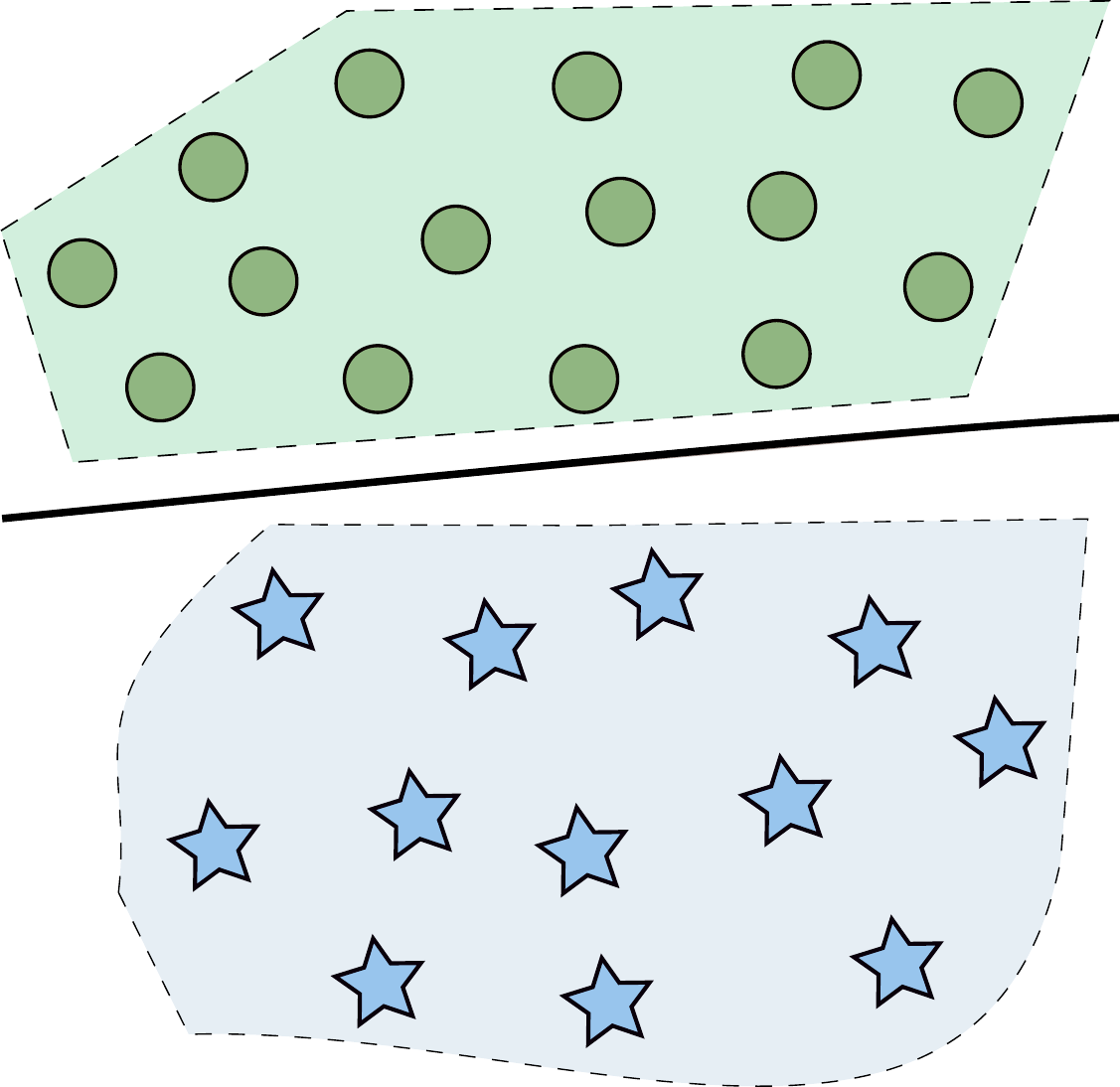}}
	\hspace{1mm}
	\subfigure[Types of drift]{\includegraphics[width=0.36\columnwidth]{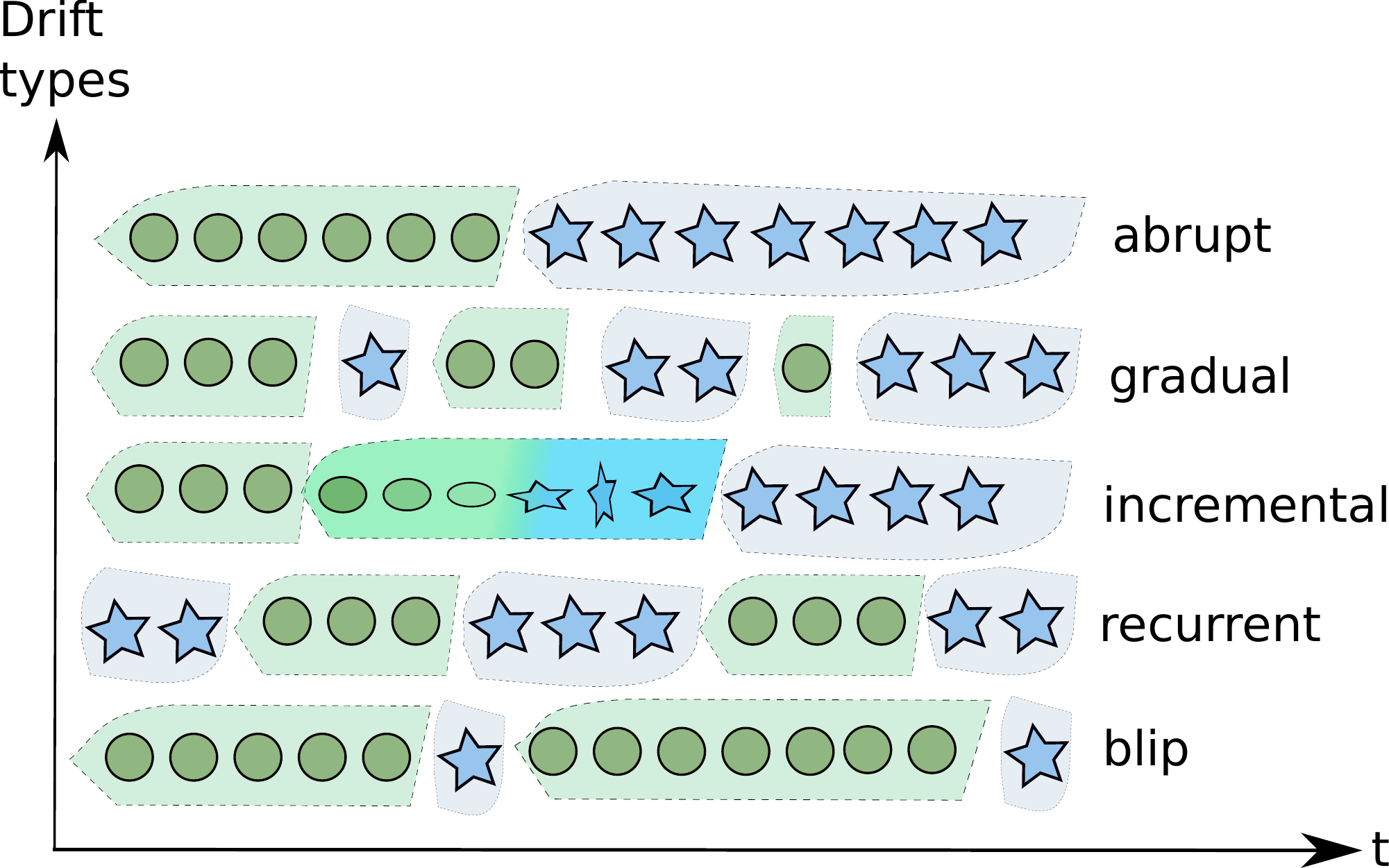}}	
	\caption{(\textit{a}) nature of drift depending on what is changing; (\textit{b}) types of drift.} 
	\label{fig:virtual_real_drift}
\end{figure}

Learning and adaptation to drift in non-stationary environments requires modeling approaches capable of monitoring, tracking and adapting to eventual changes in the produced data. In this context, a drift may occur in the feature domain (new features appear, part of them disappear, or their value range evolves) or in the class domain (new classes emerge or some of them  fade along time). More formally, concept drift between time step $ts_{0}$ and $ts_{1}$ can be defined as:
$\exists x \in X \colon p_{ts_{0}}(\textbf{x},y) \neq p_{ts_{1}}(\textbf{x},y)$, where $p_{ts_{0}}(\textbf{x},y)$ and $p_{ts_{1}}(\textbf{x},y)$ are the joint probability distributions at time steps $ts_{0}$ and $ts_{1}$, respectively, between the input variables $\textbf{x}$ that conform a data sample and the target variable $y$. Specific terminology is often used to indicate the cause or nature of changes. In terms of what is changing (see Figure \ref{fig:virtual_real_drift}), drifts can be:

\begin{description}
	\item [\textit{Real drift}] when the posterior probabilities of classes $p(y|\textbf{x})$ vary over time independently from variations in $p(\textbf{x})$, having an impact on unconditional probability density functions, and
	\item [\textit{Virtual drift}] when the distribution of the input data $p(\textbf{x})$ changes without affecting the posterior probabilities of classes $p(y|\textbf{\textbf{x}})$, but affects only the conditional probability density functions.
\end{description}

Concept drift can be also categorized in terms of speed and severity \citep*{minku2011online}. On the one hand, in the case of speed, an \textit{abrupt} drift may occur when a change happens suddenly between two classification contexts, whereas a \textit{gradual} drift represents the case when dealing with a smooth transition between two concepts. When there are several intermediate concepts in between the old and the new concept, the change is \textit{incremental}. Likewise, if previously known concepts reoccur after some time, it is considered as a \textit{recurrent} drift. Finally, a challenge may emerge when an outlier (blip) can be mixed with concept drift; in this case no adaptation is needed because it is a temporary event that does not affect the future data and thus the subsequent learning of the algorithm. Figure \ref{fig:virtual_real_drift} also shows these types of drifts. On the other hand, severity can be regarded as the amount of changes that a new concept causes; therefore, a measure of severity can be computed as the percentage of the input space whose target class has changed after the drift. It is worth mentioning that many efforts have been devoted to the characterization of concept drift \citep*{webb2016characterizing,khamassi2018discussion}, however, it is still an open issue in the state of the art due to the complexity of characterizing manifold types of data changes over time.

Algorithms that handle concept drift can be designed for adaptation or detection purposes, even both when the active strategy requires a detection mechanism to trigger the adaptation process. The adaptation process can be carried out proactively (first detecting concept drift, so that only the model gets updated when a drift is detected, known as \textit{active} or \textit{informed} approaches), or passively (updating the model continuously every time new data samples are received, known as \textit{passive} or \textit{blind} approaches). \textit{Passive} approaches are effective with gradual drifts and recurring concepts (although \textit{active} ones are also able to do it but with more difficulties), and they are more recommendable for \textit{batch learning}. \textit{Active} approaches work well when the drift is abrupt, and they are more recommendable for OL.

\begin{figure}[!ht]
	\centering
	\captionsetup{justification=centering}		
	\includegraphics[width=0.45\columnwidth]{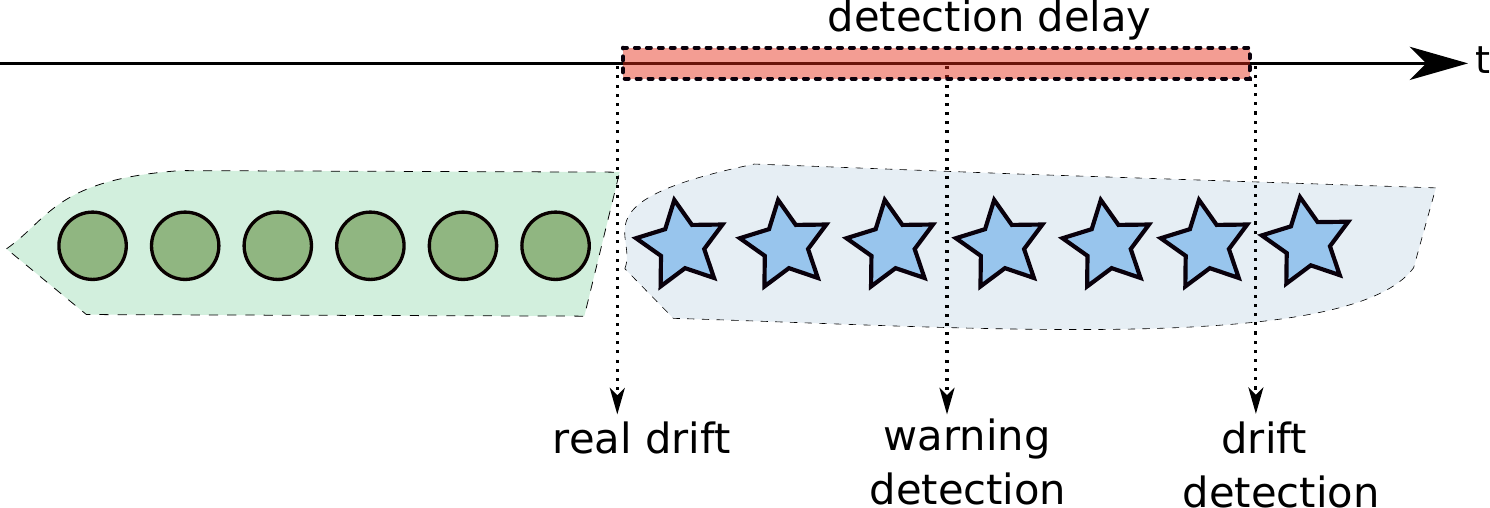}
	\caption{Drift detection example.} 
	\label{fig:drift_adap_detect}
\end{figure}

Drift detectors are methods that can detect data distributions changes based on information about a base learner performance (e.g. classification errors) or the incoming data. Such changes usually trigger the need for updating, replacing or retraining the model (or the ensemble). Drift detectors may return not only signals about drift occurrence, but also warning signals, which represent the moment when a change is suspected and a new training set representing the new concept should start being collected to retrain the models, as Figure \ref{fig:drift_adap_detect} shows. 

\subsection{Applications}

The {\em Internet of Things (IoT)} has become one of the main applications of OL, since it is generating huge quantity of data continuously in real time. The IoT may be defined as sensors and actuators connected by networks to computing systems \citep*{McKinsey}, which can monitor/manage the health and actions of connected objects/machines in real-time. It has the potential to fundamentally shift the way we interact with our surroundings, or dramatically improve health outcomes with wearable devices and portable monitors. It has also the potential to deliver trillions of dollars in economic growth in the coming years. It will boost productivity, drive the emergence of new markets, and encourage innovation. In \citep*{Gartner}, the authors predict that the IoT will grow at a $32\%$ rate from $2016$ through $2021$, reaching an installed base of $25.1$ billion units. In $2021$, $7.6$ billion ``things" will ship, with $64\%$ being consumer applications. The IoT will support total spending on endpoints and services of about $\$3.9$ trillion in $2021$. The characteristics of the IoT are:
\begin{description}
	\item[Interconnectivity:] anything can be interconnected with the global information and communication infrastructure,
	\item[Things-related services:] the IoT is capable of providing thing-related services within the constraints of things (e.g. privacy protection and semantic consistency between physical things and their associated virtual things). In order to provide thing-related services within the constraints of things, both the technologies and information in the physical world will change,
	\item[Heterogeneity:] IoT devices are heterogeneous as based on different hardware platforms and networks. They can interact with other devices or service platforms through different networks,
	\item[Dynamic changes:] the state of devices change dynamically, e.g., sleeping and waking up, connected and/or disconnected as well as the context of devices including location and speed. Moreover, the number of devices can change dynamically, and
	\item[Enormous scale:] the number of devices that need to be managed and that communicate with each other will be at least an order of magnitude larger than the devices connected to the current Internet. The ratio of communication triggered by devices as compared to communication triggered by humans will noticeably shift towards device-triggered communication. Even more critical will be the management of the data generated and their interpretation for application purposes. This relates to semantics of data, as well as efficient data handling.
\end{description}

In IoT applications, OL algorithms are needed to manage the data currently generated, at an ever increasing rate, from applications such as: sensor networks, measurements in network monitoring and traffic management, log records or click-streams in web exploring, manufacturing processes, call detail records, email, blogging, twitter posts and others \citep*{vzliobaite2016overview}. In fact, all data generated can be considered as streaming data since it is obtained in specific intervals of time.

\subsection{Available Open Software/Frameworks}
The references provided here contain software implementations for algorithms that can work on stationary and non-stationary scenarios. We do not claim this list to be exhaustive, but provides several opportunities for novices to get started, and established researchers to expand their contributions, all the while advancing the OL field by tackling some of the open challenges described in the next subsection.
\begin{description}
	\item[MOA:] is probably the most popular open source Java framework for data stream mining \citep*{MOA-Book-2018}, 
	\item[SAMOA:] is a Scalable Advanced Massive Online Analysis tool \citep*{morales2015samoa} for distributed stream learning,
	\item[Scikit-Multiflow:] is implemented in Python given its increasing popularity in the ML community \citep*{skmultiflow}, and it is inspired by MOA. It contains a collection of ML algorithms, datasets, tools, and metrics for OL evaluation,
	\item[Scikit-Learn:] although it is mainly focused on batch learning, this framework\footnote{https://scikit-learn.org/stable/} also provides researchers with some OL methods, such as Multinomial Naive Bayes, Perceptron, a Stochastic Gradient Descent classifier, a Passive Aggressive classifier, among others, and
	\item[SparkML:] is a Spark-based ML library \citep*{meng2016mllib} for large-scale data processing.
\end{description}

\subsection{Recent Challenges and Future Trends}

Next, we summarize some of the most remarkable challenges and trends in Online Learning \citep*{ramirez2017survey,gomes2017survey,wang2018systematic,onlineMLBigData}: 

\begin{description}
	\item [Structured prediction:] instead of having only one output attribute to predict, in structured prediction we may have several output attributes at the same time, that can be numeric or discrete. If they are discrete, we  refer to it using the term  {\em multi-label learning} and if they are numeric, the term  {\em multi-target learning},
	\item [Semisupervised and delayed learning:] as the number of class labels available may be small, we may need to used semi-supervised techniques to get benefit of having huge quantities of unlabeled data,  to make predictions. Another interesting and real case scenario is the one where class labels arrive with delay. We may use also semi-supervised techniques to deal with this delayed setting,
	\item [Active learning:] if there is a cost for obtaining the label of an instance,  active learning can be used to decide which instances to select to pay the cost for the label, optimizing the cost and the number of labels used,
	\item [Data preprocessing:] in high-dimensional data, using all attributes can not be feasible, and we may need to preprocess the data to perform feature selection, or feature transformation. How to do that in an efficient way is  still quite challenging, and
	\item [Imbalanced learning and  anomaly detection:] in many applications data is not balanced, the distribution of the class labels is not uniform and may be evolving other time. One application of imbalanced learning is anomaly detection, where the problem consists in predicting when an anomaly appears. As anomalies appear with a very low frequency, it is a classical example of imbalanced learning.
	\item [Distributed computation:] when dealing with large quantities of data, an important trend will be how to do OL using distributed streaming engines, such as Apache Spark, Apache Flink, Apache Storm among others. Algorithms have to be distributed in an efficient way, so that the performance of the distributed algorithms does not suffer from the network cost of distributing the data, and
	\item [The use of neural networks:] how to execute only doing one pass over the data will be an important future area of research, considering that the standard deep learning techniques currently do several passes over the data.
\end{description}

\section{Spiking Neural Networks (SNNs)}\label{SNNs_sec}

The computational power of bio-inspired systems has attracted increasing attention from research community \citep*{Kasabov2019}. Despite the lack of consensus about the information processing involved in brain, biological processes have served as reference for recent computational models. ANNs were developed as simplified versions of biological neural networks in terms of structure and function. SNNs have raised as the new generation of neural networks, a more biological realistic approach by utilizing spikes, incorporating the concepts of space and time trough neural connectivity and plasticity. They deal with precise timing spikes improving the traditional ANNs in terms of accuracy and computational power, being potentially better suited for hardware implementation due to their simple ``integrate-and-fire" nature (see Section \ref{models}). There are several trade-offs of the hardware implementations of SNNs: there are no multiplications as in traditional models, pulse processing can be implemented using shifts and adds, and interconnections transmit only a single bit instead of real numbers. Sparse and asynchronous communication can also be easily implemented. However, it should be remarked that this prospective advantage does not manifest itself yet when implementing SNNs in a general purpose computer platform. In order to have an overview of SNNs in comparison with some statistical methods (e.g. multiple linear regression, k-nearest neighbors, support vector machines, etc.) and second generation of ANNs (e.g. multilayer perceptron, convolutional neural networks, etc.), we present Table \ref{table1}.

\begin{figure}[!ht]
	\centering
	\captionsetup{justification=centering}	
	\includegraphics[width=0.6\columnwidth]{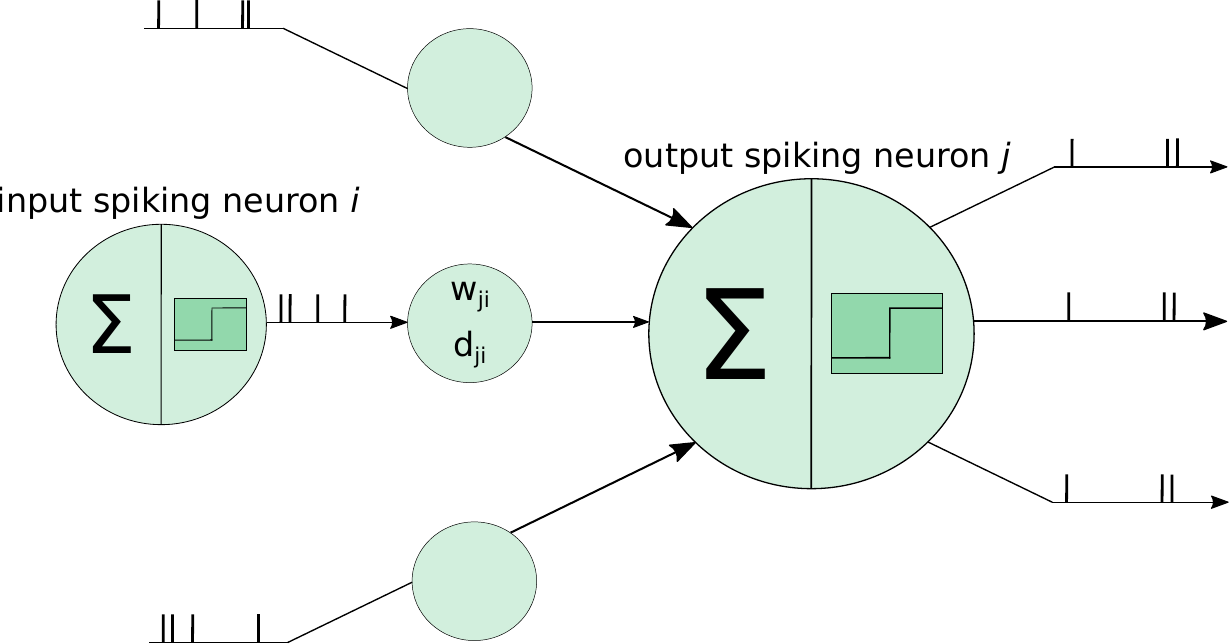}
	\caption{Scheme for SNNs.}
	\label{fig:snn}
\end{figure}

We would like to underline the importance of model interpretation. By having the possibility of building a model visualization or extracting the set of rules that govern the SNN model, we are able to interpret the internal mechanisms behind the model learning. In contrast to other ANNs where it is hard to look into the network and figure out exactly what or how it has learned, some SNNs allow us to know about their learning process \citep*{soltic2010knowledge}.

\begin{table}[!ht]
	\resizebox{\textwidth}{!}{%
		\begin{tabular}{@{}lccc@{}}
			\toprule
			\textit{} & \textbf{Statistical methods} & \textbf{2nd generation} & \textbf{SNNs} \\ \midrule
			\textit{Information representation} & Scalars & Scalars & Spike sequences \\
			\textit{Input data representation} & Scalars, Vectors & Scalars, Vectors & Whole SSTD patterns \\
			\textit{Learning} & Statistical, limited & Hebbian rule & Spike-time dependent \\
			\textit{Dealing with SSTD} & Limited & Moderate & Excellent \\
			\textit{Parallelisation of computations} & Limited & Moderate & Massive \\
			\textit{Hardware support} & Standard & VLSI & Neuromorphic VLSI \\ \bottomrule
		\end{tabular}%
	}
	\caption{Comparison of SNNs with statistical methods and second generation of ANNs. SSTD refers to Spatio- and Spectro-Temporal Data, and VLSI to Very Large Scale Integrated.}
	\label{table1}
\end{table}

Before finishing this subsection, some problems of SNNs are presented. They highly depend on the optimization of a large number of parameters, they show an unknown behavior for different types of spatio-temporal data, and they suffer from the lack of a solid consensus about the best information encoding scheme and neuron model. Despite these limitations, they are one of the most promising technique in ML. Now the scope is placed on SNNs (see Figure \ref{fig:snn}), providing a general overview on this family of connectionist models, showing their principles, current applications, and future trends and challenges \citep*{nikola2018time}.

\subsection{Biological Inspiration}

\begin{figure}[!ht]
	\centering
	\captionsetup{justification=centering}	
	\includegraphics[width=\columnwidth]{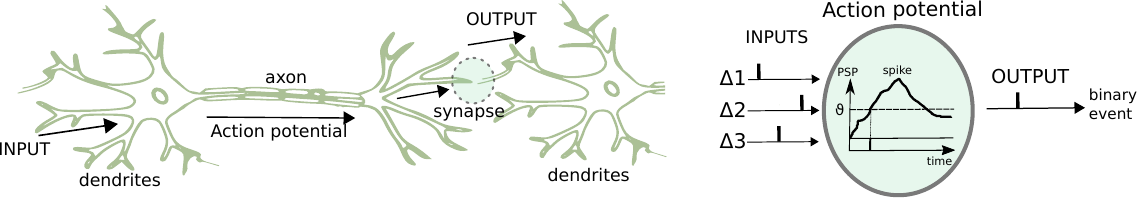}
	\caption{Biological neuron and its association with an artificial spiking neuron \citep*{gerstner2002spiking}.}
	\label{fig:bio_motivation}
\end{figure}

Brain-inspired neural networks show similarities with respect to the way brains process information, they are intrinsically associated with the computation of neuronal units that use spikes. The utilization of spikes brings together the definitions of time varying post-synaptic potential (PSP), firing threshold ($\vartheta$), and spike latencies ($\Delta$), as depicted in Figure \ref{fig:bio_motivation}. They try to simulate the processes carried out between the neurons (synapses) in a network.

\subsection{Data and Information Representation as Spikes}

\begin{figure}[!ht]
	\centering
	\captionsetup{justification=centering}	
	\includegraphics[width=\columnwidth]{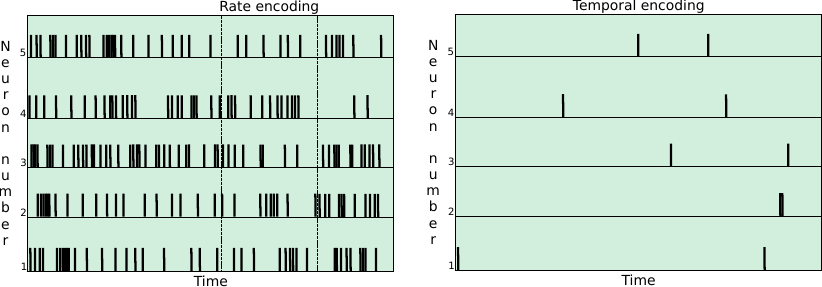}
	\caption{\textit{Temporal encoding} versus \textit{Rate-based encoding}.}
	\label{fig:rate_vs_temp}
\end{figure}

Before presenting the input data to the SNN, it must be encoded into spike trains in order to apply the neuron model. This encoding part aims to generate spiking patterns that represent the input stimuli, and it is still an open issue in neuroscience (what is the information contained in such a spatio-temporal pattern of spikes?, what is the code used by neurons to transmit that information?, how might other neurons decode the signal?, etc.), but traditionally it has been shown that most of the relevant information is contained in the mean firing rate of neurons. In the literature we can find two main encoding schemes: \textit{temporal encoding} and \textit{rate-based encoding} (see Figure \ref{fig:rate_vs_temp}). The first one is used over the latter one when patterns within the encoding window provide information about the stimulus that cannot be obtained from spike count. The \textit{rate-based encoding} scheme is based on a spiking characteristic within a time interval (e.g. frequency), in the \textit{temporal encoding} scheme the information is encoded in the time of spikes.

\textit{Rate-based encoding} schemes (``rate as a spike count", ``rate as a spike density", and ``rate as a population activity") correspond to three different notions of mean firing rate (either an average over time, or an average over several repetitions of the experiment, or an average over a population of neurons respectively). \textit{Temporal encoding} schemes are based on spike timing: ``time-to-first-spike" (when a code for the timing of the first spike contains all information about the new stimulus), ``phase" (when we can apply a ``time-to-first-spike" encoding scheme also where the reference signal is not a single event, but a periodic signal), and ``correlations and synchrony" (where we use spikes from other neurons as the reference signal for a spike code).  

\subsection{Spiking Neuron Models}\label{models}

\begin{figure}[!ht]
	\centering
	\captionsetup{justification=centering}		
	\subfigure[LIF model]{\includegraphics[width=0.50\columnwidth]{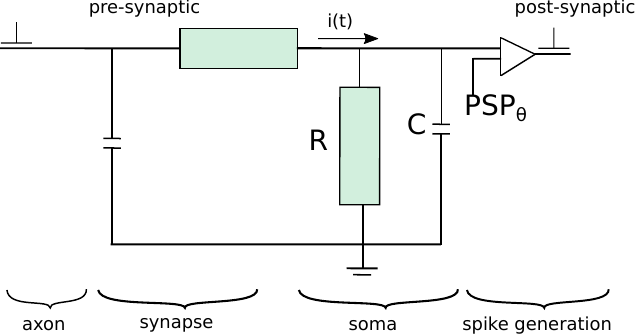}}
	\hspace{10mm}
	\vspace{5mm}
	\subfigure[Izhikevich model]{\includegraphics[width=0.40\columnwidth]{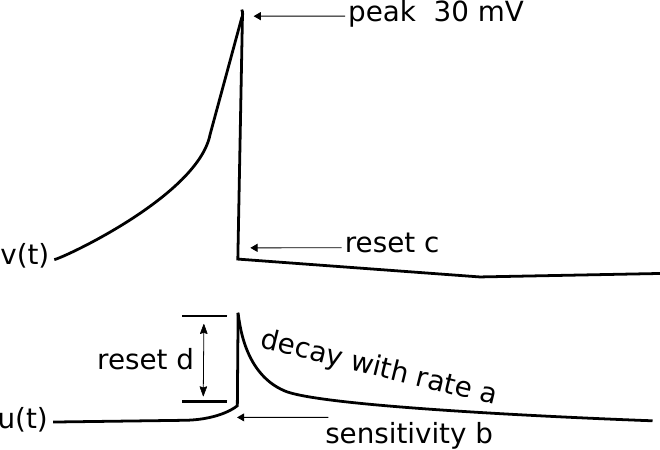}}
	\hspace{10mm}
	\subfigure[Hodgkin-Huxley model]{\includegraphics[width=0.70\columnwidth]{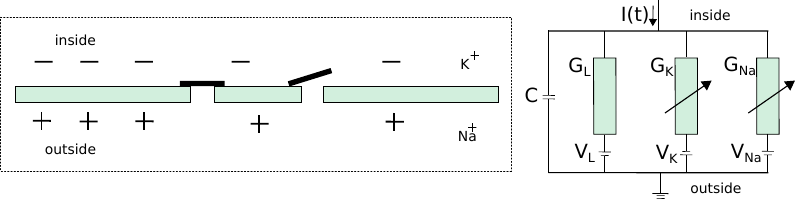}}
	\caption{Three of the most representative neuron models.} 
	\label{fig:models}
\end{figure}

A spiking neuron model is a mathematical description of the properties of neurons that generate electrical potentials across their cell membrane. Some of the most relevant neuron models are described subsequently, and graphically represented in Figure \ref{fig:models}.

\begin{description}
	\item [\textit{Leaky Integrate-and-Fire (LIF)}  \citep*{gerstner2002spiking}] model, where a neuron is considered as an electrical circuit and the current potential is computed with an appropriate equation. Model consists of capacitor \textit{C} in parallel with resistor \textit{R}, driven by a current $I(t) = I_{R}+ I_{cap}$. The standard form of the model is defined as $\tau_{m}\frac{du}{dt}=-u(t)+RI(t)$, where $\tau_{m}=RC$ is the membrane time constant. Here spikes are events characterized by a firing time $t^{f}:u(t^{f})=\vartheta$, and after $t^{f}$, the potential is reset to a resting potential $u_{r}$. In a more general form, the LIF model can also include a refractory period, in which the dynamics are interrupted for an absolute time $\Delta^{abs}$. LIF model is simple and computationally effective, and it is the most widely used spiking neuron model despite other more biologically realistic models. One of the most used approximations of the LIF model is the Spike Response Model (SRM), where each time a neuron receives an input from a	previous neuron, its internal state (membrane potential) changes. In summary, a neuron emits an spike each time its membrane potential reaches a threshold value ($\theta$) (see Figure \ref{fig:bio_motivation}). Then, after emitting  the spike, the neuron goes through a phase of high hyperpolarization during which it is impossible to emit a second spike for some time (refractory period).
	\item [\textit{Hodgkin-Huxley} \citep*{hodgkin1952quantitative}] model, where a semipermeable cell membrane separates the interior of the cell from the extracellular liquid, acting as a capacitor ($C$). When an input current $I(t)$ is injected into the cell, it may add further charge on $C$, or leak through the channels in the cell membrane. Because of active ion transport through the cell membrane, the ion concentration inside the cell is different from that in the extracellular liquid. The Nernst potential generated by the difference in ion concentration is represented by a battery. A detailed description of the influences of the conductance of three ion channels ($Na$, $K$ and $L$) on the spike activity of the giant axon of squid is given by the following equation: $\sum_{ch}i_{ch}(t)=G_{Na} \cdot m^{3} \cdot h \cdot (v_{c}-v_{Na}) + G_{K} \cdot n^{4} \cdot (v_{c}-v_{K}) \cdot G_{L} \cdot (v_{c}-v_{L})$, and its differential equations: $\dfrac{dm}{dt}=\alpha_{m}(v_{c}) \cdot (1-m)-\beta_{m}(v_{C}) \cdot m$, $\dfrac{dn}{dt}=\alpha_{n}(v_{c}) \cdot (1-n)-\beta_{n}(v_{C}) \cdot n$, and $\dfrac{dh}{dt}=\alpha_{h}(v_{c}) \cdot (1-h)-\beta_{h}(v_{C}) \cdot h$. In the equations, $G_{Na}$, $G_{K}$, $G_{L}$ are the conductance of the sodium, potasium, and leakage channels; $V_{Na}$, $V_{K}$, and $V_{L}$ are constants called reverse potentials; $m$ and $n$ control the $Na$ channel and variable $h$ controls the $K$ channel; $\alpha$ and $\beta$ are empirical functions of $v_{c}$.	
	\item [\textit{Izhikevich} \citep*{izhikevich2007dynamical}] model claims to be as biological plausible as the Hodgkin-Huxley model with computational efficiency of LIF models. It is defined by the equation $v'=0.004v^{2}+5v+140-u+I$ and $u'=a(bv-u)$. Here it is considered that $if v\geq30mV, then v \longleftarrow c, u \longleftarrow u+d$; where $a, b, c$ and $d$ are parameters of the model, $v$ represents the membrane potential, and $u$ the membrane recovery.
	
	\item [\textit{Probabilistic} \citep*{kasabov2010spike}] model, which stores its information in connection weights and probabilistic parameters related to spikes to the occurrence and the propagation of spikes.
	
\end{description}

\subsection{Learning Methods and Algorithms in SNN}\label{learning_methods}

Synaptic plasticity is the capacity of synaptic connections to change their strength, which is the basis of the learning and memory processes in biological neural networks. Several synaptic plasticities coexist, mainly differing on the time scale and the conditions required for the induction. Regarding the time scale, some of them decay on the order of about $10-100$ ms, while others, such as Long-Term Potentiation (LTP) or Long-Term Depression (LTD), persist during a longer time. Referred to the conditions for the induction, some synaptic plasticities depend only on the history of pre-synaptic stimulation (independently of the post-synaptic response), while others depend on the coincidence of pre-synaptic and post-synaptic activity, on the temporal order of their spikes, and on other factors (e.g. a concentration of specific chemicals).

In supervised learning, Supervised Hebbian Learning (SHL) rule provides probably the most straightforward solution for from a biologically realistic view. Here, a spike-based Hebbian process is supervised by an additional ``teaching" signal that reinforces the post-synaptic neuron to fire at
the target times and to remain silent at other times. ReSuMe \citep*{ponulak2005resume} and SpikeProp \citep*{bohte2002error} are two of the most representative algorithms in supervised learning for SNN. In unsupervised learning, by modifying synaptic strengths of Hebbian processes, the connections are reorganized within a neural network and, under certain conditions, may lead to an emergence of new functions (input clustering, pattern recognition, source separation, dimensionality reduction, formation of associative memories or self-organizing maps). It was demonstrated that the change in the synaptic efficacy after several repetitions of the experiment was a function of the relative differences of the spike times; and that whereas pre-synaptic spikes that precede post-synaptic ones induce potentiation, the reversed order of spikes induce synaptic depression. This phenomenon is called Spike-Time Dependent Plasticity (STDP) \citep*{bi2001synaptic}.

\subsection{Applications}

Many practical applications of SNNs can be found in the fields of motor control (e.g. \citep*{dewolf2016spiking} for the motor control of the arm), robotics control based \citep*{bing2018survey}, trajectory tracking, decision making with application to financial market, spatial navigation and path planning, decision making and action selection, rehabilitation, image and odor recognition, spatial navigation and mental exploration of the environment, etc. \citep*{ponulak2011introduction}.

It is worth mentioning that there is a great opportunity for the use of SNNs in the \textit{Green Artificial Intelligence} paradigm \citep*{villani2018meaningful}, which is becoming an important subfield of Artificial Intelligence. It is based on the reduction of the usage of resources while training and testing our models, using techniques of energy efficiency, energy aware computing, hardware accelerators, and embedded systems. Here, the use of SNN in OL allows for a very fast real-time simulations of large networks and a low computational cost.

\subsection{Available Hardware and Software/Frameworks}

Again, we do not claim this subsection to be an entire list, but offers some of the most known hardware/software/platforms for beginners to get started. Now we reference some of these software implementations/frameworks for SNNs.

\begin{description}
	\item[Brian:] is an open source simulator for SNNs \citep*{goodman2008brian} written in Python\footnote{http://briansimulator.org/}, 
	\item[Cypress:] is a C++ SNNs simulation framework\footnote{https://github.com/hbp-unibi/cypress} that provides a wrapper around PyNN, and allows to directly run networks on the Human Brain Project \citep*{markram2012human} neuromorphic hardware systems,
	\item[Neuron:] is a simulation environment \citep*{carnevale2006neuron} that can perform efficient discrete event simulations of SNNs with LIF models, as well as hybrid simulations involving both LIF neurons and cells with voltage−gated conductances. It is especially recommended for simulations that involve cells with complex anatomical and biophysical properties,
	\item[Nest:] is a simulation environment \citep*{van2018performance} best suited for models that focus on the dynamics, size, and structure of neural systems rather than on the detailed morphological and biophysical properties of individual neurons,
	\item[PyNN:] is a Python package\footnote{http://neuralensemble.org/PyNN/} for simulator-independent specification of neuronal network models. Once the model has been implemented with PyNN API and Python, we can run it without modification on any simulator that PyNN supports (Neuron, Nest, Brian) or on the supported neuromorphic hardware systems, such as SpiNNaker and BrainScaleS (see below),
	\item[NeuCube:] is a 3D eSNN computational neurogenetic model \citep*{kasabov2014neucube} to map, learn and mine brain data, 
	\item[PCSIM:] is a software package written in Python (although its computational core is written in C++) for simulation of neural circuits \citep*{pecevski2009pcsim} primarily designed for distributed simulation of large scale networks of SNNs, and 
	\item[ANNarchy:] is a neural simulator \citep*{vitay2015annarchy} designed for distributed rate-coded or SNNs (or both). The core of the library is written in C++, but it provides an interface in Python for the definition of the networks.  
\end{description}

Next, we list two of the most known available hardware platforms:

\begin{description}
	\item[SpiNNaker:] is a massively parallel computing platform \citep*{furber2014spinnaker} mainly targeted towards neuroscience, robotics, and computer science. It is based on numerical models running in real time on custom digital multicore chips using the ARM architecture. Next generation small scale test chips of the SpiNNaker architecture is available for first test users since early $2018$,
	\item[BrainScaleS:] is a project\footnote{https://brainscales.kip.uni-heidelberg.de/index.html} that aims at understanding and emulating function and interaction of multiple spatial and temporal scales in brain information processing. It is based on physical (analogue or mixed-signal) emulations of neuron, synapse and plasticity models with digital connectivity, running up to $10$ thousand times faster than real time. Next generation small scale test chips of the BrainScaleS architecture is available for first test users since early $2018$.
\end{description}

\subsection{Recent Challenges and Future Trends}

Previous subsections have been underlined that SNNs are biologically more realistic than traditional ANNs, and computationally more powerful. The question that arises here is why community still works with continuous neural networks instead of the theoretically superior SNNs. Now, we present some of the most relevant challenges for the wide use of SNNs:

\begin{description}
	\item [Which model to apply?:] there is no unified framework or an agreed upon base neuron model of SNNs yet. Different models are often applied, which makes tough the task of carrying out insightful comparisons. Besides, it is well known the trade-off between the biological plausibility and the computational cost of the model: the model choice ranges from LIF (simple but efficient, they are usually chosen by computer scientists and engineers) to the Hodgkin-Huxley (sophisticated but slow, they are usually chosen by neuroscience researchers).
	\item [Information encoding:] Although it is already known that brains do not work with real numbers but with timed spikes, the crux of the issue is how information is encoded with such spikes, being one of the big unresolved challenges of neuroscience. Here we find again a lack of agreement on what constitutes a good encoding scheme. 
	\item [Learning based on spike timing:] how to achieve a good learning algorithm is still one of the biggest challenges of SNNs, mainly motivated by two reasons. On the one hand, the biological learning method of brains is not still well understood, so it is hard to imitate it with artificial learning methods. On the other hand, the discontinuous nature of spikes makes difficult the design of calculus based approaches.
	
\end{description}

\section{SNNs in Online Learning Scenarios}\label{SNNs_OL}

Data streams may exhibit temporal dependencies between class labels, which formally occur whenever the current sample label $y^{t}$ is influenced by previous sample labels ($y^{t-1}$, $y^{t-2}$,...). Temporal dependence can help to determine how the input features relate with each other over time. SNNs leverage spike information representation so as to build spike-time learning rules that have shown the ability to capture temporal associations between temporal variables in streaming data. Additionally, STDP and Hebbian learning are biologically plausible local learning rules in SNN models; the use of some SNNs (e.g. eSNNs) in OL allows for a very fast real-time and reducing the computational complexity of the learning process, given its locality which lends itself well for parallel implementation. In terms of adaptation to the drift, most off-the-shelf classification models need to be retrained if they are used in changing environments, and fail to scale properly. Some SNNs can overcome this drawback, e.g. the evolving nature of eSNNs (based on the merging process of similar neurons) makes possible the accumulation of knowledge as data become available, without the requirement of storing and retraining the model with past samples. Finally, they have also shown to be very competitive as concept drift detectors \citep*{lobo2018drift}.  

\subsection{Existing SNN Approaches for Online Learning: Drawbacks and Trade-offs}

SpikeProp \citep*{bohte2002error} is a learning rule  based on gradient descent for training SNNs, and it is able to solve complex classification problems. However it presents several drawbacks: it tends to be trapped in local minima, its convergence is not guaranteed because depends on fine parameter tuning before starting, it is too slow to be used in an online setting, and the large number of synaptic connections makes difficult to scale up when a high dimensional dataset is considered.

The SHL rule, already mentioned in Section \ref{learning_methods}, exhibits two main drawbacks. On the one hand, as during training ``teaching" signal currents suppress all undesired fires, the only correlations of pre-synaptic and post-synaptic activities happen around the target firing times. In other occasions, this correlation is not present, and there is no mechanism to weaken these synaptic weights that make the neuron to fire at undesired times during the testing phase. On the other hand, synapses change their weights even when the neuron fires already exactly at the desired times. Therefore, SHL can achieve stable solutions only by adding additional constraints or more learning rules. These problems are resolved in ReSuMe \citep*{ponulak2005resume}, where an instructive signal modulates synaptic plasticity not to have a marginal direct effect on the post-synaptic somatic membrane potential. Despite this method was claimed to be suitable for OL, the network structure used is fixed and does not adapt to incoming stimuli. Finally, SHL and ReSuMe are suitable for training in single-layer networks, and it is more recommendable the use of multi-layer feedforward or recurrent neural networks for many other tasks because they are capable of performing more complex computation than single-layer networks \citep*{ponulak2011introduction}.

More recently, SpikeTemp method proposed in \citep*{wang2017spiketemp} offers an enhanced rank-order-based learning method for SNNs with an adaptive structure where the precise times of incoming spikes are used to determine the required change in synaptic weights.

However, most of them are unable to predict inputs after just one presentation of the training samples, and then other studies have tackled OL in a more realistic approach. One of the most promising SNNs for OL is the Evolving Spiking Neural Networks (eSNNs) \citep*{schliebs2013evolving}, based on the Thorpe model \citep*{thorpe2001spike} which is a simplified version of LIF model (it simplifies the leaky operation of the computational neuron), and where its neural model allows for a very fast real-time simulation of large networks and a low computational cost. These properties make them suitable for these scenarios, where stringent constraints on computational cost and processing time prevail. In addition, their evolving nature (spiking neurons evolves incrementally over time to infer temporal patterns from data) allows to accumulate knowledge as data arrive, without storing and retraining the model with past data. They are also recommendable for non-stationary environments, because changes in the input stream data are encoded immediately as binary events or spikes, which is one of the most suitable data encoding strategies for adapting to drifts. Recently, eSNN was modified in \citep*{lobo2018} to improve their adaptation to the drift, and to consider OL in a more realistic form by limiting the size of the neuron repository, avoiding its incremental growing, which is unfeasible in OL scenarios. They were also used for first time as drift detector in \citep*{lobo2018drift}, where it takes advantage of their neurons merging mechanism by using it as detector. To date, with these few exceptions, there is a lack of efficient and scalable SNN-based algorithms for OL scenarios.

\subsection{Topics of Future Interest}

Next, we summarize some of the most interesting topics for the future in the field of SNNs and OL: 

\begin{description}
	
	\item [\textit{Lifelong Machine Learning} (LML):] it involves the capability of the model to smoothly update its captured knowledge to take simultaneously into account different tasks and data distributions, yet still being able to reuse and retain captured knowledge and skills recurrently occurring over time \citep*{chen2016lifelong}. Hence, LML is a paradigm requiring notably higher time scales where data (and tasks) become available only during certain periods time and probably with no access to previously seen data. Thereby, it is imperative to build on top of previously learned knowledge, from where the connection between LML and OL emerges. 
	\item [Deep SNN learning:] future research should addresses deep learning of spatio-temporal patterns from streaming data, deep knowledge representation and its online adaptation \citep*{nikola2018time}.
	\item [Human-computer interaction:] a particularly interesting field for future application of SNNs in OL is of human-computer interaction where biological spiking neurons need to communicate with software in a potentially very rapid manner.
	\item [Others:] parameter optimization in evolving scenarios, visualization of SNN models, spatial mapping of input variables into a SNN architecture, neuromorphic hardware systems for real-time applications, etc., should be tackled in a near future.
	
\end{description}

\section{Conclusions}\label{conc}

In this work we have analyzed the OL and SNNs fields from an introductory perspective to serve as an entry point for the application of SNNs to OL, which is a very hot topic in the research community due to the large number of real applications based on stream data, even more in those scenarios where data is affected by non-stationary events, provoking the so-called concept drift. SNNs are considered the third generation of neural networks, and have revealed themselves as one of the most successful approaches to model the behavior and learning potential of the brain, allowing for a very fast real-time simulation of large networks and a low computational cost. They have also shown a very good behavior in drift detection and drift adaptation situations, often present in OL scenarios. All of this leads us to consider both fields in an incredibly interesting intersection. Still, much progress has to be made in both fields for tackling their respective open challenges, but we should be aware of the importance of merging OL and SNNs in order to solve real problems with the computational power of these bio-inspired systems.

\section*{Acknowledgements}\label{acknow}

This work was supported by the EU project \textit{iDev40}. This project has received funding from the ECSEL Joint Undertaking (JU) under grant agreement No 783163. The JU receives support from the European Union's Horizon 2020 research and innovation programme and Austria, Germany, Belgium, Italy, Spain, Romania. It has also been supported by the Basque Government (Spain) through the project \textit{VIRTUAL} (KK-2018/00096).

\section*{Bibliography}

\bibliographystyle{model5-names}
\bibliography{biblio}

\end{document}